\documentclass[sn-vancouver, iicol]{sn-jnl}
\usepackage{natbib}
\usepackage{amsmath}

\jyear{2021}%

\theoremstyle{thmstyleone}%
\theoremstyle{thmstyletwo}%

\theoremstyle{thmstylethree}%

\begin{document}

\title{Regulated Pure Pursuit for Robot Path Tracking}

\author*[1]{\fnm{Steve} \sur{Macenski}}\email{s.macenski@samsung.com}
\author[2]{\fnm{Shrijit} \sur{Singh}}\email{shrijit.singh@learner.manipal.edu}
\author[3]{\fnm{Francisco} \sur{Mart\'in}}\email{francisco.rico@urjc.es}
\author[3]{\fnm{Jonatan} \sur{Gin\'es}}\email{jonatan.gines@urjc.es}

\affil*[1]{\orgdiv{R\&D Innovations}, \orgname{Samsung Research America}, \orgaddress{\street{Clyde Ave}, \city{Mountain View}, \postcode{94043}, \state{CA}, \country{United States}}}

\affil[2]{\orgdiv{Department of Computer Science}, \orgname{Manipal Institute of Technology}, \orgaddress{\street{Udupi - Karkala Rd}, \city{Eshwar Nagar, Manipal}, \postcode{576104}, \state{Karnataka}, \country{India}}}

\affil[3]{\orgdiv{Intelligent Robotics Lab}, \orgname{Rey Juan Carlos University}, \orgaddress{\street{Camino del Molino}, \city{Fuenlabrada}, \postcode{28943}, \state{Madrid}, \country{Spain}}}


\abstract{The accelerated deployment of service robots have spawned a number of algorithm variations to better handle real-world conditions.
Many local trajectory planning techniques have been deployed on practical robot systems successfully.
While most formulations of Dynamic Window Approach and Model Predictive Control can progress along paths and optimize for additional criteria, the use of pure path tracking algorithms is still commonplace.
Decades later, Pure Pursuit and its variants continues to be one of the most commonly utilized classes of local trajectory planners.
However, few Pure Pursuit variants have been proposed with schema for variable linear velocities - they either assume a constant velocity or fails to address the point at all.
This paper presents a variant of Pure Pursuit designed with additional heuristics to regulate linear velocities, built atop the existing Adaptive variant.
The \textit{Regulated Pure Pursuit algorithm} makes incremental improvements on state of the art by adjusting linear velocities with particular focus on safety in constrained and partially observable spaces commonly negotiated by deployed robots. 

We present experiments with the Regulated Pure Pursuit algorithm on industrial-grade service robots.
We also provide a high-quality reference implementation that is freely included ROS 2 Nav2 framework at \url{https://github.com/ros-planning/navigation2} for fast evaluation.}


\maketitle

\section{Introduction}
\label{sec:introduction}
Dynamic Window Approach (DWA)~\cite{dwa}, Pure Pursuit (PP)~\cite{pure_pursuit}, and Model Predictive Control (MPC)~\cite{mpc} are by far the most commonly deployed path trackers. 
They all have a strong heritage for reliability in a wide range of environmental conditions.
DWA and MPC are often, but not always, formulated as multi-objective trajectory generation problems to maximize criteria such as avoiding dynamic obstacle collisions on top of path tracking.
This has made them particularly well suited for many robotics applications where dynamic robot behaviors are rewarded.
A great deal of work has been conducted on these allowing them to be found on many commercially available robots today.

However, there still exist many applications of deployed robot systems where this can be considered a detractor.
Surveyed among high-end research and service robot navigation systems, pure path tracking continues to be a common theme in many robots \cite{genora,se2_nav}.
Among single-objective path trackers, a simple and reliable method continues to be exploited decades after its development: Pure Pursuit. 

Pure Pursuit uses simple geometry to find the curvature of a path required to drive a robot towards a given point on the path.
The algorithm itself does not place any stability restrictions on the translational velocities during operation, however it also lacks any schema for selecting them.
Near-universally, implementations use a fixed speed. 
Many variations of Pure Pursuit exist, however most address the more obvious area of the selection of lookahead points, which largely aids in stabilizing convergence behaviors towards the path at a wider range of velocities. 
The Pure Pursuit algorithm was not developed with service and industrial robots in mind, which have additional safety requirements making it further unrealistic to move at a fixed velocity. 

This work proposes an incremental improvement on the Pure Pursuit path tracking algorithm by describing a reference method of adjusting translational and rotational velocities to improve safety and operability in a broad range of common deployed robot applications.
We improve the Adaptive Pure Pursuit (APP) algorithm \cite{MITDARPA} by regulating velocities via penalizing sharp changes in path curvature and proximity to obstacles - two of the most common events requiring conscientious navigation behaviors.
The \textit{Regulated Pure Pursuit} (RPP) algorithm slows sharp turns into partially observable dynamic environments (aisles, hallways, intersections) to reduce the likelihood and impact of potential collisions.
It also reduces linear velocity in close proximity to obstacles such as people and fixed infrastructure to reduce likelihood of collision in constrained indoor environments.
Finally, it additionally includes preemptive collision detection missing from other variations.

Another main contribution of this work is in describing and providing free and high quality implementations of PP, APP, and RPP for evaluation.
It is integrated into the Nav2 mobile robot navigation system commonly used by researchers and adds a new capability to the framework \cite{nav2}.
This work is well documented, tested, and is in use on robots deployed today.

This paper is organized as follows. First, in section \ref{sec:related_work}, we will describe the conventional Pure Pursuit Algorithm, providing a mathematical formulation 
that will allow us to describe related works that provide variations to the original algorithm. In section \ref{sec:description}, we describe the 
Regulated Pure Pursuit Algorithm, our main contribution to this paper. Section \ref{sec:implementation} will describe its implementation in a real robotic system, and in section 
\ref{sec:experiments}, we will experimentally validate the improvement introduced with our contribution. Finally, we will provide the conclusions in section \ref{sec:conclusion}.

\section{Related Work}
\label{sec:related_work}

Pure Pursuit~\cite{MITDARPA} is a widely used algorithm for path tracking. It is simple but effective.
Considering a path $\mathcal{P}$ as an ordered list of points  $\mathcal{P} = \{p_0, p_1, ..., p_n\}$ where $p_i = (x_i, y_i) \in \mathcal{P}$.
A local trajectory planner is a function $f$ that determines the linear and angular velocity to track a reference path $\mathcal{P}_t$ at time $t$ (Eq. \ref{eq:controller}).

\begin{equation}\label{eq:controller}
(v_t, \omega_t) = f(\mathcal{P}_t)
\end{equation}
\hspace{8pt}

\begin{figure}[ht]
    \centering
    \includegraphics[width=0.35\textwidth]{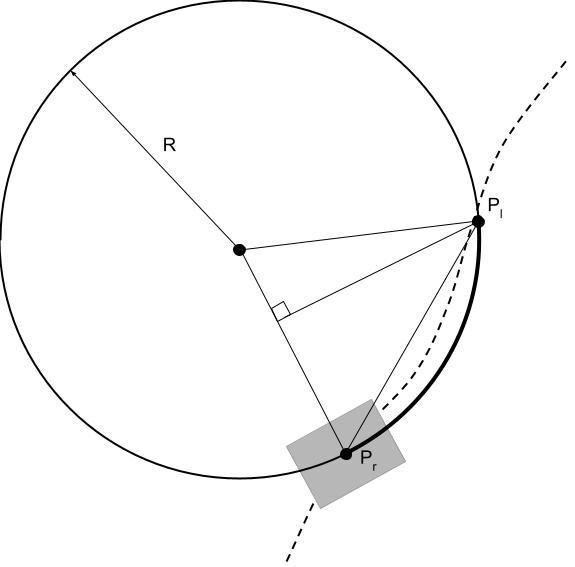}
    \caption{Geometry of finding the path curvature.}
    \label{fig:lookahead_curve}
\end{figure}

Figure \ref{fig:lookahead_curve} shows visual geometric representations of Pure Pursuit.
First, it determines the closest point $p_r$ on $\mathcal{P}_t$ to the robot position.
Using a given lookahead distance, $L$, the lookahead point $p_l$ is determined as the first $p_i$ at least $L$
distance away from $p_r$ in Eq. \ref{eq:lookahead}.

\begin{equation*}
dist(p_i) = \sqrt{(x_r - x_i)^2 + (y_r - y_i)^2}\\
\end{equation*}
\begin{equation}\label{eq:lookahead}
p_l =  p_i \in \mathcal{P}_t,  \left\{ \begin{array}{l} 
dist(p_{i-1}) < L \\
dist(p_{i}) \ge L \\
\end{array} \right. 
\end{equation}
\hspace{8pt}

With a known $p_l$, we can determine curvature of the circle (recall, $R = 1 / \kappa$) using simple geometry.
If $\mathcal{P}_t$ is represented in vehicle base coordinates, $\mathcal{P'}_t$, where the robot position is the origin, then the curvature can be represented as

\begin{equation}
  \label{eq:t}
    \kappa = \frac{2 \; y'_l}{L^2} \\
\end{equation}
\hspace{8pt}

where $\kappa$ is the path curvature required to drive the robot from its starting position to the lookahead carrot, $y'_l$ is the lateral coordinate of the 
lookahead point $p'_l$, and $L$ is the desired distance between $p_r$ and $p_l$. 


Figure \ref{fig:lookahead_curve} shows this visually, where $L$ is represented geometrically as the circle's chord.
With curvature to drive towards the lookahead point, the commands can be send to a robot controller.
This process is then updated at the desired rate. 


The primary parameters of the Pure Pursuit path tracker are simply the velocity of travel $v_t$ and the distance $L$ along the path used to select the lookahead point $p_l$.
In the standard formulation, this lookahead point $p_l$ is a distance $L$ from the robot tuned to achieve an acceptable trade-off between oscillations centered around the path (shorter distances) and slower convergence (longer distances).
There exists a broad range of admissible lookahead distances \cite{pure_pursuit}.

PP and its variants are applicable to Ackermann and differential-drive robots due to PP's formulation supporting dynamics with longitudinal motion and a turning rate in body-fixed frame.
For Ackermann robots to use PP, the path must be feasible given the kinematic-constraints of the platform - while differential-drive robots may follow any holonomic path using PP.
While omni-directional robots may utilize PP variants, they will be restricted from performing lateral movements.

There are several known downsides of this approach.
In high curvature situations, Pure Pursuit is known to have overshoot or undershoot behaviors resulting in path deviations, even in a well tuned system \cite{pure_pursuit}.
This is typically not a major concern for autonomous driving applications which naturally has a minimum turning radius limit, but a more substantive issue for smaller-scale applications like industrial and consumer robots.
It also does not specify any translational velocity specification criteria during execution.
While this can be beneficial as it allows for a great deal of flexibility with different linear velocity profiles, in practice, without described methods, nearly all known variants use a constant translational velocity profile.
This is an unsafe defacto-standard. 




Variants of this algorithm have been proposed to increase path tracking stability by varying computations of the lookahead point.
MIT's entry into the DARPA Urban Challenge implemented the \textit{Adaptive} Pure Pursuit (APP) algorithm for lane following while varying the lookahead distances proportionally to the translational velocity \cite{MITDARPA}.
For velocities in the operating range, a mapping of lookahead distances to velocities is required such that they have an acceptable trade-off between oscillation and slower convergence to the path. 
A common formulation for this is in Eq. \ref{eq:app}, where $L_t$ is the lookahead distance, $v_t$ is the translational velocity, and $l_{t}$ is a lookahead gain representing the amount of time to project $v_t$ forward \cite{Interpolation}.

\begin{equation}\label{eq:app}
L_t = v_t l_{t}
\end{equation}
\hspace{8pt}

A recent variant has been created to significantly improve path tracking accuracy of Pure Pursuit \cite{new_pp}.
This variant also modifies the process of determining the lookahead point, but instead adjusts this point off the path to address the long-standing edge case of short-cutting during path tracking in substantial turns. 
While it also contains a heuristic change to the translational velocity during that edge-case, its policy is derived from on-road properties and ackermann kinematics that does not generalize to other applications outside of on-road driving of Ackermann vehicles. 
It does not solve the problems facing deployment of PP techniques to general mobile robotics applications for which this work focuses (e.g. practicable translational velocity control); though its improvements in path tracking accuracies are notable. 

Pure Pursuit nor its variations account for dynamic effects of the vehicle. 
As PP is geometrically derived, vehicle dynamics are not modeled in this style of path tracking algorithm.
There are other types of path trackers that do account for vehicle dynamics \cite{survey}.
This paper uses PP and its variants for comparison to offer a candid contrast with the same or directly analog parameters to clearly indicate the differences in performance due to the contribution \footnote{Not leaving the opportunity for misleading results by comparing unrelated families of methods with unknown degrees of tuning}.

\section{Regulated Pure Pursuit}
\label{sec:description}

The Regulated Pure Pursuit algorithm is designed for service and industrial mobile robots in real-world constrained and partially observable environments.
It provides methods for adapting the robot's translational velocity to current conditions; robot systems cannot simply barrel through aisles and facilities at full speed without regard.
These methods are linear regulation cost functions that provide high-quality behavior across a variety of practical mobile robot environments derived from commonplace requirements on reducing robot velocities in the presence of sharp turns or when operating in confined regions.

The first phase of the Regulated Pure Pursuit algorithm is to transform the input path $\mathcal{P}_t$  $\{p_0, ..., p_j, p_r, ..., p_n\}$ into the robot's base coordinate frame $\mathcal{P'}_t$ and prune it.
In doing so, determining the curvature of the path is reduced to the simple algebraic expression shown in Eq. \ref{eq:t}.
Before transformation, $p_r$ is determined and all prior points $\{p_0, ..., p_j\}$ are permanently pruned from the stored path to prevent unnecessary future transformation of obsolete data.
The transformed path ($\{p'_r, ..., p'_n\}$) is also pruned for all $p_i$ points where $dist(p'_r, p'_i) >> L_t$ as they are sufficiently far away that they will never need to be considered at $t$.
These far path points continue to be held in the stored path for future iterations, until a new path is received, as the robot progresses along the path.

Next, RPP will utilize the same lookahead selection mechanics as Adaptive Pure Pursuit described in Eq. \ref{eq:app}.
The lookahead distance $L_t$ is thusly proportional to the speed $v_t$ and a lookahead gain $l_t$ such that longer distances are used while moving faster.
This stabilizes the path tracking behavior over larger ranges of translational velocities \cite{adaptive}.
This distance is used to select the lookahead point $p_l$. 
While interpolating between path points was found to demonstrably improve smoothness on sparse paths at autonomous vehicle speeds, empirically this did not contribute much benefit at the service and industrial robot speeds using typical grid map planning resolutions (0.025 m - 0.1 m) \cite{Interpolation}.
However, interpolation is beneficial for use in sparser path resolutions (0.1 - 1.0m).

The desired linear velocity, $v_{t}$, is next further processed by the curvature and proximity heuristics.
Both heuristics are applied to linear velocity and we take the maximum of the two.

The purpose of the \textbf{curvature heuristic} is to slow the robot to $v'_t$ during sharp turns into partially observable environments, such as when entering or exiting hallways and aisles commonly found in retail, warehouses, factories, schools, and shopping malls.
This allows for significantly safer traversal when making blind turns.
This heuristic is applied to the linear velocity $v_t$ when the change in curvature $\theta$ is above a minimum threshold $T_\theta$.
This minimum radius restricts velocity scaling in minor turns or path variations that do not require slowing.
The curvature velocity $v'_t$ returned by this heuristic is selected as: 


\begin{equation}
  \label{eq:b¡a}
  v'_t = 
 \begin{cases} 
      v_{t} &  \kappa > T_\kappa, \\
      \frac{v_{t}}{r_{min} \; \kappa} & \kappa \le T_\kappa \\
   \end{cases}
\end{equation}

Where $r_{min}$ is the minimum radius to apply the heuristic. 
This formulation is a mathematical reduction of a simple error calculation between the minimum radius and the radius of a circle represented by $\kappa$ and can be trivially derived.

The \textbf{proximity heuristic} is applied to the linear velocity $v_t$ when the robot becomes in close proximity to dynamic obstacles or fixed infrastructure. 
The purpose of this is to slow the robot when in constrained environments where the potential for collision is particularly high.
Reducing the speed near fixed infrastructure lowers the likelihood of collision by lessening the impact of small path variations in tight spaces. 
Lowering the speed of industrial and service robots in close proximity to dynamic agents, such as humans, is a common safety requirement- allowing a robot to reactively stop faster to prevent potential injury. 
The linear formulation of this heuristic reduces the speed by the ratio of $d_O / d_{prox}$ with a gain, $\alpha$, to adjust the response for individual systems.
The other formulations tested, such as exponential and quadratic, far too significantly penalized proximity to objects and contained only a narrow band of gains which would result in an acceptable trade-off between proximity to obstacles and velocities to accomplish the robotic task.
This linear heuristic in Eq. \ref{eq:b} has a broad range of $\alpha$ which may be finely tuned by a system designer and is derivative of the well-used Adaptive Pure Pursuit's formulation \cite{adaptive}.

\begin{equation}
  \label{eq:b}
  v'_{t} = 
 \begin{cases} 
      v_{t} \; \frac{\alpha \; d_{O}}{d_{prox}} & d_{O} \leq d_{prox} \\
      v_{t} & d_{O} > d_{prox} \\
   \end{cases}
\end{equation}

Where $d_{prox}$ is the proximity distance to obstacles to apply the heuristic, $d_{O}$ is the current distance to an obstacle, and $\alpha$ is a gain to scale the heuristic function for aggressive behavior, with the requirement that $\alpha \leq 1.0$.
A higher $\alpha$ lowers the velocity of the robot in proximity to obstacles more expeditiously.
The value of $d_{prox}$ should be established based on the system requirements of a robot's application for how close an obstacle can be before the robot begins slowing its maximum velocity.

After velocity regulation, the algorithm then determines the path curvature using Eq. 1.
The angular velocity is computed using the regulated velocity, not desired linear velocity, which prevents consequential undershoot behavior relative to target curvatures \cite{pure_pursuit}.
Finally, the angular velocity, $\omega_t$, is then trivially found as Eq. \ref{eq:final}.

\begin{equation}
  \label{eq:final}
\omega_t = v'_{t} \; \kappa
\end{equation}
\hspace{8pt}

The final step of the algorithm is to check our path tracking command for current or imminent collisions, new to RPP.
A given angular velocity $\omega_t$ and regulated linear velocity $v'_t$ can be projected forward in time, resulting in a circular arc.
Points on the arc are sampled at the grid map cell resolution forward for a set duration.
Collision checking is done based on a duration to collision rather to the lookahead point such that the robot is always, at minimum, a set duration from collision.
At slow speeds, it may not be sensible to collision check to a lookahead point tens of meters, or hundreds of seconds, away.
Rather, a temporal schema allows for fine maneuvers in confined spaces where the current velocity commands may not be admissible a short distance - but long time - away.

This controller is known to converge from the stability analysis completed in \cite{Ollero95}.
Ollera's work on the general Pure Pursuit algorithm applies to Regulated Pure Pursuit as well, showing that straight paths and constant curvature paths are stable.
Pure Pursuit itself is independent of velocity in its formulation.
It is only used in the final step to convert the path curvature into a rotational velocity for a base to track. 
The regulation heuristics proposed in this variant do not change the basis of stability of the root algorithm.

\section{Implementation}
\label{sec:implementation}
Another contribution of this work is a high-quality reference implementation of Regulated Pure Pursuit.
This reference implementation is available as one of the default path tracking algorithm plugins in the new and improved ROS 2 Navigation System, known as Nav2. 
Nav2 is a scalable navigation framework with multiple algorithm implementations, documentation, and support for building modern and reliable research and commercial navigation systems \cite{nav2}.
This optimized C++ implementation has 92\% unit test coverage and is used in the experiments in Section \ref{sec:experiments}.
It also contained a few additional features that are highlighted in this section.

Rather than creating distinct implementations of Pure Pursuit, Adaptive Pure Pursuit, and Regulated Pure Pursuit, all three have been built into a single reference implementation parameterized to enable each specific behavior.
This allows us to test these algorithms easily and analyze their run-time performance more closely knowing that they share the vast majority of their  computations. 
This also allows researchers and developers to quickly evaluate and tune these features by simply changing a handful of parameters from the available binaries.

Additionally, there is a setting to allow the robot to slow its speed when approaching the target goal pose.
This feature allows the robot to come to a more gradual stop.
When the robot is within close proximity to the goal, the translational velocity is lowered proportional to the remaining distance, up to a minimum viable velocity to make progress.

\begin{figure}[ht]
    \centering
    \includegraphics[width=0.35\textwidth]{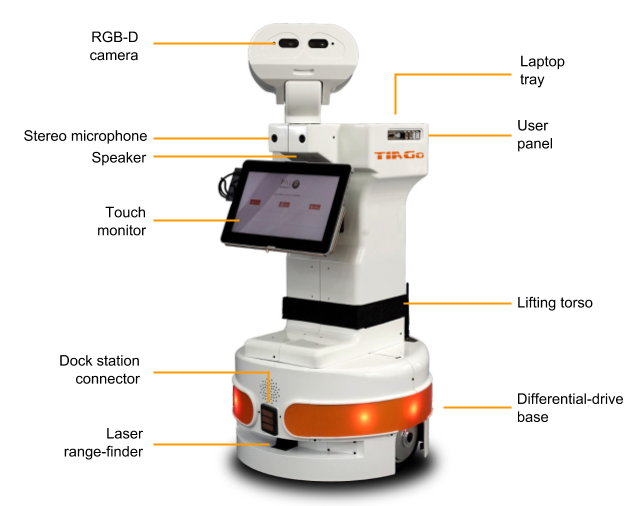}
    \caption{The Pal Robotics Tiago used in the experiments \cite{tiago_img}.}
    \label{fig:tiago}
\end{figure}

This work is specifically targeted at industrial and service robots, many of which are differential drive.
While this is of no technical concern to Pure Pursuit, many differential drive robots will utilize holonomic search-based path planners (such as Dijkstra's or A*) rather than kinematically feasible planners frequently used by Ackermann vehicles \cite{brian}.
This is of concern to the Regulated Pure Pursuit implementation as the relative heading of the path may not align with the robot's starting heading.
In this case, it is beneficial for the robot to be able to rotate to a rough starting heading before beginning to track the holonomic path.
Further, rotating to the final heading when the robot is within the translational goal tolerances is useful when being deployed in orientation-sensitive applications.
These are non-issues when using feasible planning algorithms that, no matter the drivetrain.
However, it is useful to consider this situation since holonomic grid search algorithms are popular.

Further, it is a common approach for local trajectory algorithms to leverage a rolling environmental model with the robot at its center.
The predictive collision detection algorithm will check for collisions some \textit{N} seconds in the future given the robot's current commanded speed.
If a near-future collision is detected, the robot is stopped. 
When a robot is traveling at high speeds, this may actually lapse outside of the bounds of a rolling environmental model if not carefully tuned. 
The implementation offers protections from such situations and issues stern recommendations to prevent the serious safety violation.

Users are able to tune all of the parameters related to PP, APP, and RPP.
In particular, the heuristics have parameters $r_{min}$, $\alpha$, and $d_{prox}$. 
It is possible for these to be poorly tuned such that a robot would drop to an impractically low speed when in close proximity to an obstacle or in a turn.
To prevent the robot from traveling \textit{too} slow, a minimum speed threshold is exposed such that the robot's speed will never be set below a parameterized value.
Note: the parameters and speeds in the experiment were set up such that this did not impact our experimental results in Section \ref{sec:experiments}.

An open-source community contributed feature made available is support for reversing - not only driving forward.
When a path contains a cusp or discontinuity in direction, the controllers will reverse from forward to back and vis-a-via.
When paired with a planner that can generate such paths, such as the Hybrid-A* planner in Nav2, the robot can track more complex paths.

\section{Experiments and Analysis}
\label{sec:experiments}

This section describes the experiments to demonstrate the benefits of our contribution compared to other existing Pure Pursuit variants.
One experiment was conducted in simulation and three others were conducted with a physical Tiago robot (Figure \ref{fig:tiago}).
Tiago is used in industrial and service applications and the test environment is shown in Figure \ref{fig:exp1}, corresponding to a university campus building.




In all three hardware experiments below, the same navigation configurations are utilized.
The maximum linear speed $v_{max}$ is set to $0.8 m/s$, the maximum acceleration $a_{max}$ is $0.2 m/s^2$, and the maximum angular speed $\omega_{max}$ is $3.2 rad/s^2$.
The lookahead distance $L_t$ is set between $0.25 m$ and $1.2 m$ and lookahead time is $1.0 s$ for Adaptive and Regulated Pure Pursuit; while Pure Pursuit uses the maximum of $1.2 m$.
In all four experiments, no replanning was utilized such that the distance to path data is regular and can be meaningfully analyzed.
SLAM Toolbox was used to generate the 2D map and AMCL was used to localize within it \cite{slam, nav2}.

\subsection{Path Tracking Experiment}
\begin{figure}[ht]
    \centering
    \includegraphics[width=0.45\textwidth]{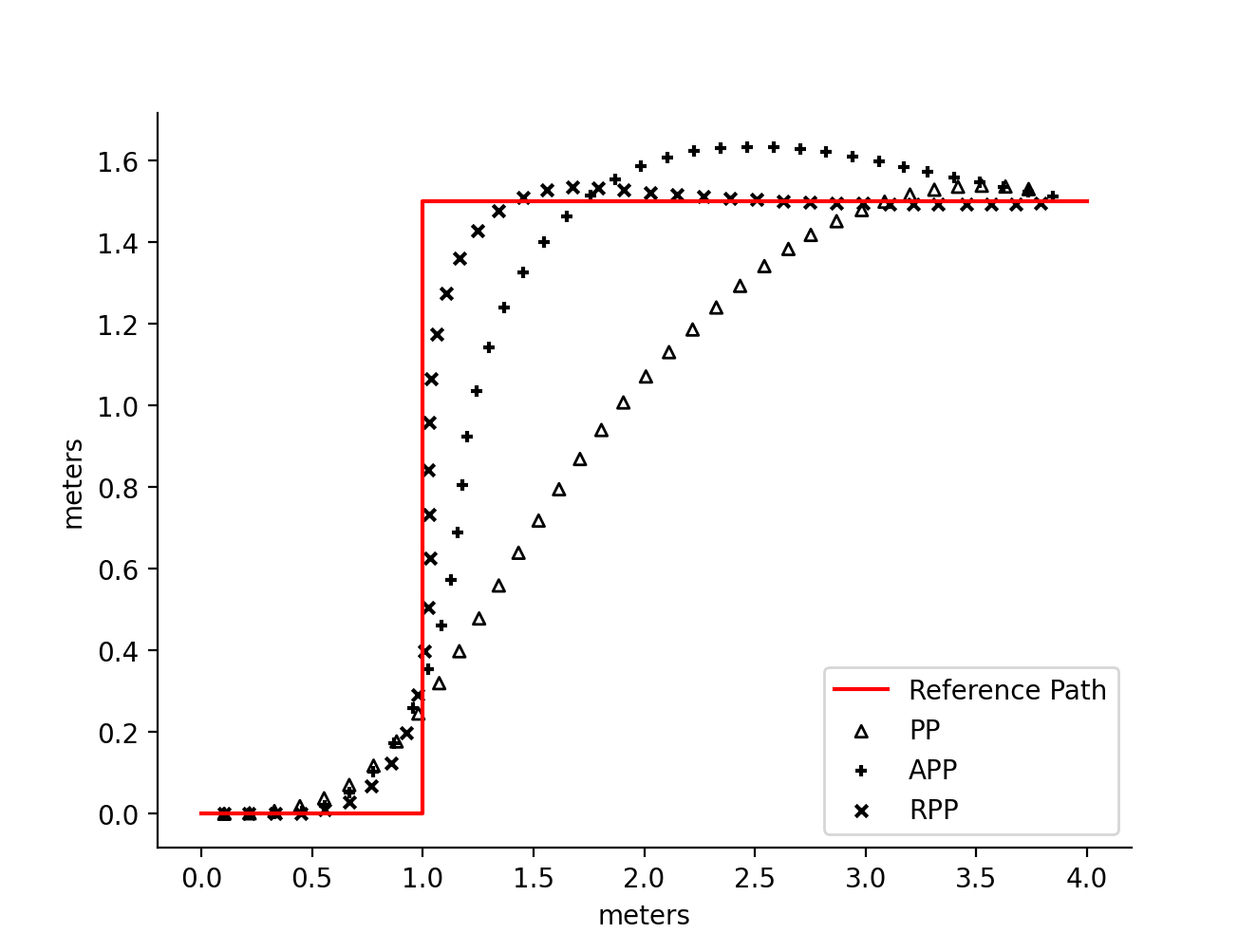}
    \caption{Comparison of variants of Pure Pursuit.}
    \label{fig:tracking_comparison}
\end{figure}

\begin{figure}[ht]
    \centering
    \includegraphics[width=0.45\textwidth]{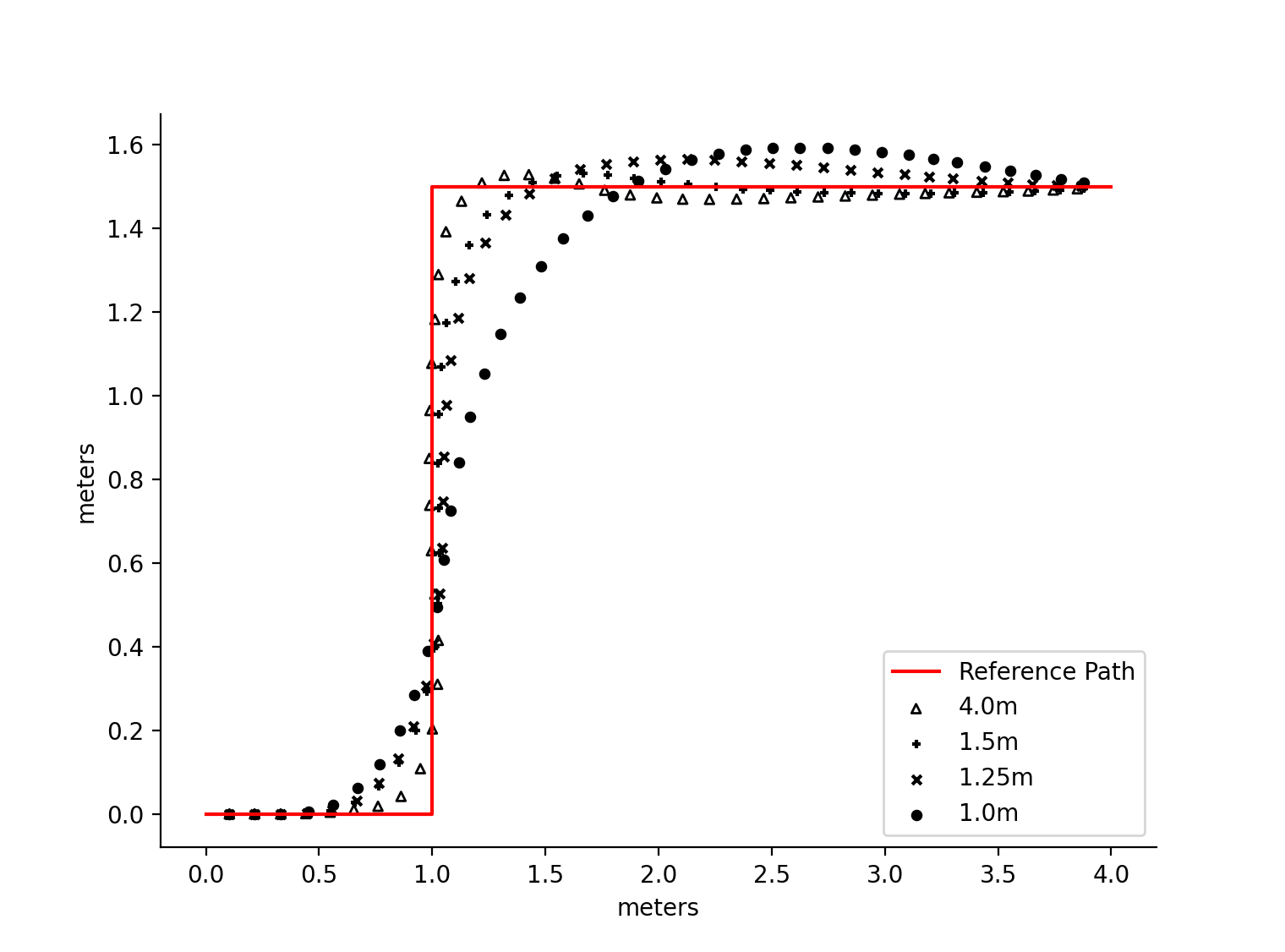}
    \caption{Effects of varying $r_{min}$ in Regulated Pure Pursuit.}
    \label{fig:reg_variables}
\end{figure}

A simulation experiment was conducted to analyze the improvements Regulated Pure Pursuit can offer during sharp turns in ideal conditions.
This simulation was conducted using the Turtlebot 3 robot, the Gazebo simulator, and an empty environment with ground truth information made available to remove the contribution of odometric and localization error.
The robot followed the reference step function path indicated by the red piece-wise line in Figures \ref{fig:tracking_comparison} and \ref{fig:reg_variables} using the three variants of Pure Pursuit.

Figure \ref{fig:tracking_comparison} displays the comparison with the different methods.
The same parameters were utilized in this experiment as the hardware experiments, with the exception of the desired linear velocity, set to $1.0 m/s$ and the curvature heuristic minimum radius, set to $1.5 m$.
The overshoot observed during sharp turns was notably lower in Regulated Pure Pursuit due to the controlled slow down during sharp changes in curvature allowing the robot to track the path more finely.
The slow down also triggers the use of a closer adaptive lookahead point further increasing local stability. 
The mean tracking error from the reference path of Regulated Pure Pursuit was merely $0.03 m$.
The others experienced over an order of magnitude difference at $0.10 m$ and $0.19 m$ for APP and PP, respectively.
While the main contribution of this work is \textit{not} an improvement on path tracking, it is a convenient emergent property to further ensure safety. 
This experiment showcases that the algorithm can follow a computed collision-free path in confined spaces more closely and thusly more safely. 

Using the ideal environment, the effects of varying the curvature heuristic's minimum radius were also studied.
From Figure \ref{fig:reg_variables}, it can be observed that larger values of $r_{min}$ allow a robot to follow the reference path more closely.
This is due to the curvature triggering reactions quickly on approach to the discontinuity.
When $r_{min}$ was increased from $1.0 m$ to $1.5 m$, the mean tracking error decreases from $0.8 m$ to $0.3 m$.
This reduction in tracking error does come at a cost: a reduction in the speed resulting in longer navigation times.
For this experiment, when $r_{min} > 1.5 m$, the improvement on tracking error became insignificant, but navigation times begin to spike as the speed during turns trends towards the minimum allowed speed.

\subsection{Blind Turning Experiment}
\begin{figure}[ht]
    \centering
    \includegraphics[width=0.25\textwidth]{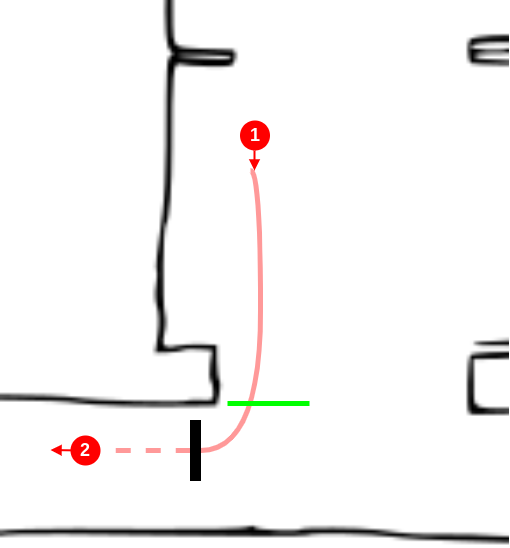}
    \caption{A blind corner experiment with an obstacle just out of view. When the robot cross the green line, the obstacle is set.}
    \label{fig:exp2}
\end{figure}

This experiment evaluates the three algorithms' performance when the robot encounters an unexpected obstacle that blocks its path around a blind corner.
The experiment consists of the robot making a sharp 90-degree turn (Figure~\ref{fig:exp2}) behind which there is an obstacle. 
The obstacle is such that the robot cannot observe it before beginning to execute the curve and has minimal time to react, placed after the  robot crosses the green line.

For each algorithm, the experiment is repeated ten times.
The purpose of this experiment is to test the reaction of each algorithm in a common but particularly dangerous situation. 
While PP and APP do not themselves contain collision detection capabilities, for a fair comparison, each are evaluated using the collision checking method described in Section \ref{sec:description}.

\begin{table}[ht]
\caption{\label{tab:exp2}Result of the Blind Turning Experiment.}
\centering
\begin{tabular}{|c|c|c|c|}
\hline                   & \textbf{PP} & \textbf{APP} & \textbf{RPP} \\\hline 
Avg Stopped Distance ($m$) & $0.15$     &    $0.16$    &  $0.24$     \\\hline 
\end{tabular}
\end{table}

Table \ref{tab:exp2} shows the results of this experiment; in all cases, none of the methods resulted in a collision.
This shows that the collision detection features of RPP (applied to all methods) are sufficient to prevent blind collisions at service robot speeds.
A similar experiment without RPP's collision detection yielded collisions in nearly all instances.
The RPP algorithm displayed a 33\% increase in average stopped distance relative to APP.
APP and PP, both which do not change their speeds in turns, had expectedly similar behavior.
This is an improvement in reaction time for safe navigation in environments with many dynamic agents potentially cornering at the same time.
While the $8 cm$ increase does not seem like much, with many robots navigating independently for their own tasks, this additional precaution is the difference between robots scraping by and colliding with permanent damage.
This highlights that RPP's slowing of the robot in blind turns helps to reduce risk in partially observable settings. 

\subsection{Confined Corridor Experiment}

\begin{figure}[ht]
    \centering
    \includegraphics[width=0.40\textwidth]{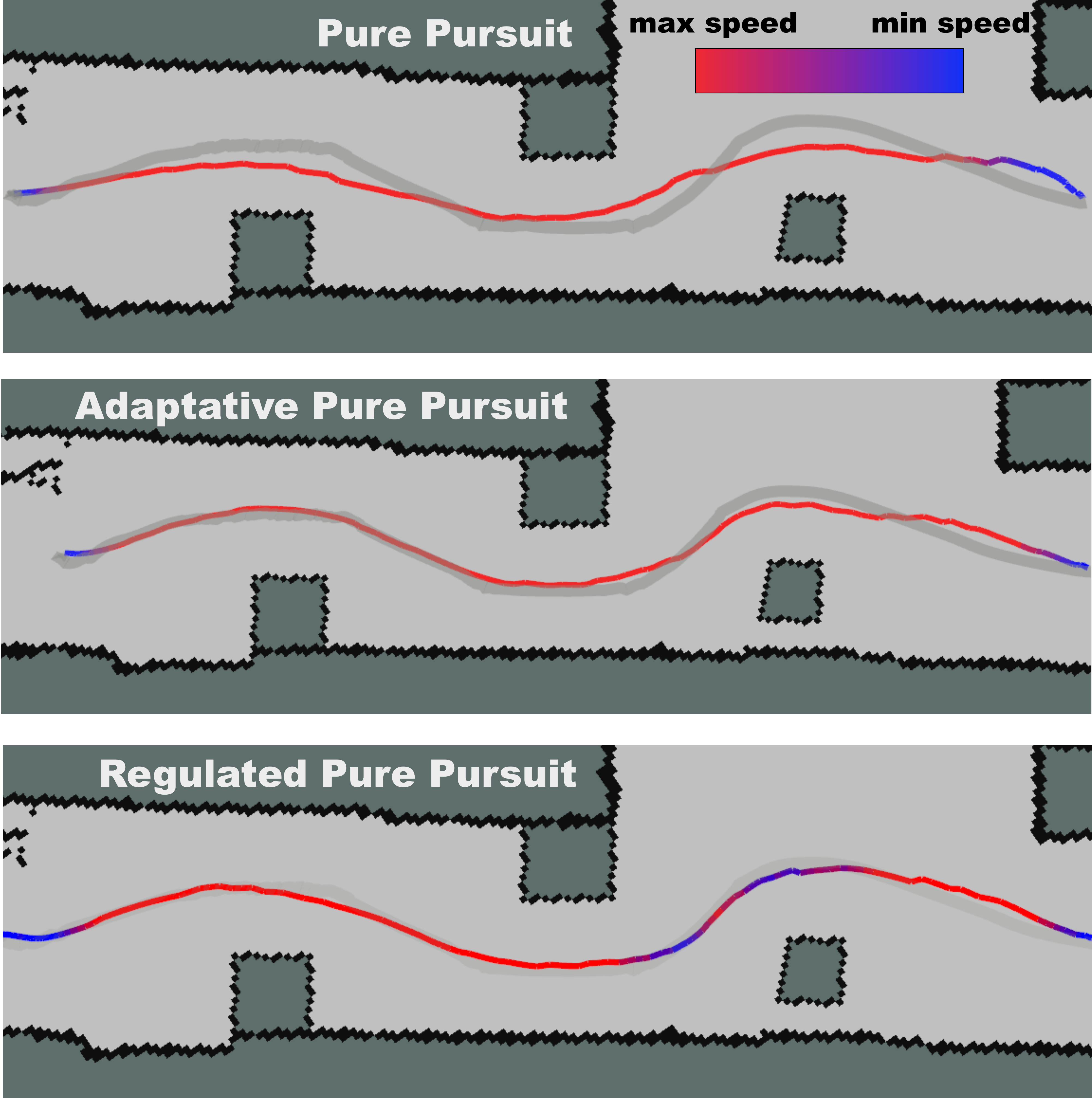} 
    \caption{Calculated path in gray and the robot's traveled path, colored by speed, during the corridor experiment.} 
    \label{fig:curve-exp1}
\end{figure}

The confined corridor experiment was designed to evaluate how well the proposed approach improves performance in narrow spaces at high speeds when making continuous changes in direction.
Figure \ref{fig:curve-exp1} shows the test environment map, creating a "slalom" pattern, starting from the left.
The total width of the corridor is 1.5 meters and the obstacles each have dimensions of roughly 0.7 m.
Each algorithm was tested five times and the data from this experiment is shown in Table \ref{tab:exp3}.

\begin{table}[ht]
\caption{\label{tab:exp3}Result of the confined corridor experiment.}
\centering
\begin{tabular}{|c|c|c|c|}
\hline                   & \textbf{PP} & \textbf{APP} & \textbf{RPP} \\\hline 
Time ($s$) & $13.0$     &    $12.7$    &  $13.3$     \\\hline 
Distance ($m$) & $9.04$     &    $9.54$    &  $9.63$     \\\hline 
Collisions & $0$     &    $0$    &  $0$     \\\hline 
Average Speed ($m/s$) & $0.679$     &    $0.736$    &  $0.682$     \\\hline 
Average Distance Obstacle ($m$) & $0.662$     &    $0.681$    &  $0.683$     \\\hline 
Average Distance to Path ($m$) & $0.100$     &    $0.059$    &  $0.052$     \\\hline 
\end{tabular}
\end{table}

Figure \ref{fig:curve-exp1} displays the robot position data overlaying the trajectory each algorithm was asked to follow.
The most telling region of this experiment to analyze is the final and sharpest turn.
RPP's proximity and curvature heuristics contributed in this location and resulted in a reduced speed and 14\% less path tracking error than APP.
The other algorithms had path undershoot due to the sharp turn and came into much closer proximity to the obstacle and wouldn't be considered functionally safe.
This undershoot was corrected in RPP by reacting to the increased curvature partially due to the deviation from the path, compounded by its approaching proximity to the obstacle, slowing the robot down.
This allowed RPP to better recover from a minor deviation off the path and navigated safer distance away from the obstacle.  
The impact of the regulation heuristics can be seen in blue as the robot weaves around the final obstacle after quickly changing direction.

The distance traveled by PP is lower than both variants due to the short cutting it displayed throughout the trajectory in Figure \ref{fig:curve-exp1}.
This habitual undershoot also led to an decreased average distance to an obstacle - passing much closer to potential collisions. 

\subsection{Full-system experiment}

\begin{figure}[ht]
    \centering
    \includegraphics[width=0.25\textwidth,angle=270,origin=c]{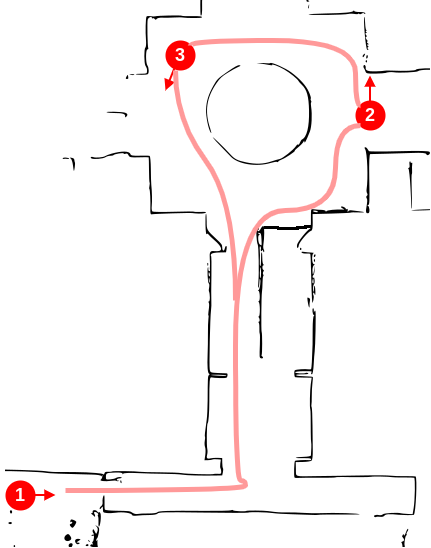} 
    \caption{Itinerary for full-system experiment. The total length is 70 meters.} 
    \label{fig:exp1}
\end{figure}

In the final experiment, the robot will follow a route through the three control points shown in Figure \ref{fig:exp1} in a campus building.
The route contains collision-free open space, confined halls, and blind sharp turn of approximately 70 m.
The robot repeats the path three times with each of the algorithms, the data is provided in Table \ref{tab:exp1}.
The purpose of this experiment is to compare the RPP algorithm's general behaviors at a system-level to the existing variants.

The data indicates little difference in overall system-level behavior - the average time to navigate was between 85-95s.
Consistent with the previous experiment, PP does display slightly shorter distance traveled and an increased tracking error - corresponding to path short-cutting.

\begin{table}[ht]
\caption{\label{tab:exp1}Result of the full-system experiment.}
\centering
\begin{tabular}{|c|c|c|c|}
\hline                   & \textbf{PP} & \textbf{APP} & \textbf{RPP} \\\hline 
Time ($s$) & $85.6$     &    $94.3$    &  $88.6$     \\\hline 
Distance ($m$) & $58.24$     &    $59.26$    &  $58.64$     \\\hline 
Collisions & $0$     &    $0$    &  $0$     \\\hline 
Average Speed ($m/s$)& $0.661$     &    $0.675$    &  $0.646$     \\\hline 
Min Distance to Obstacle ($m$) & $1.139$     &    $1.133$    &  $1.135$     \\\hline 
Average Distance to Path ($m$) & $0.062$     &    $0.049$    &  $0.043$     \\\hline 
\end{tabular}
\end{table}

The most compelling attribute of this experiment is the lack of particularly unique outcomes between the algorithms.
Although RPP slows the robot in narrow corridors and while completing sharp turns, surprisingly, these maneuvers did not meaningfully impact the high-level navigation metrics.
Time to task completion and average speeds were consistent across all three algorithms when considering the standard deviations between the trials.
The system-level performance of RPP is so similar to that of APP, we conclude that the additional benefits of RPP come at little disadvantage to a system designer.
In fact, it is possible for a solution architect to increase the maximum speed of the robot modestly when using RPP due to its slowing in turns and in proximity to obstacles (the common limiting factors of speed in robotics applications).
This meaningfully improves the overall efficiency of a robot system while securing higher-quality safety features. 
\footnote{A video with the experiments with the real robot can be found at \url{https://youtu.be/LQAzzJ8GmS0}}.

\section{Limitations}

Regulated Pure Pursuit, and other Pure Pursuit variants, suffer from a lack of modeling of the vehicle.
As this class of technique is purely geometric based on the path, it will not consider dynamic constraints while changing velocities to track the path.
Further, the global path must be feasibly drivable for a given robot platform since these methods compute velocities in the absence of kinematic limitations.  
For differential-drive robots, this may be any path.
However for Ackermann steering robots, this path must already be drivable considering the kinematic limitations on the minimum possible turning radius.

Regulated Pure Pursuit also continues to short-cut sections of paths with high curvature turns, though to a much reduced degree than PP or APP.
The work introduced in \cite{new_pp} provides an alternative method for selecting the lookahead point with the goal of more accurately tracking the reference path.
While this work also introduces a velocity heuristic that is not suitable for our class of applications, a variation on the method for selecting lookahead points has merit to be applied in mobile robotics.

\section{Conclusion}
\label{sec:conclusion}

Regulated Pure Pursuit builds incrementally on Adaptive Pure Pursuit with a focus on service robots.
This method contains a schema for selecting velocities that improves functional safety for real-world deployed robot applications.

Our demonstrations on an industrial robot exhibited improved safety around blind turns and in confined settings without significant system-level changes.
A high-quality implementation of Regulated Pure Pursuit is freely available at \url{https://github.com/ros-planning/navigation2} and is in use on robots today.

\section*{Conflict of Interest and Acknowledge}
The authors have no conflicts of interest to declare that are relevant to the content of this article. 

This work has been partially funded by Ministerio de Econom\'ia and Competitividad of the
Kingdom of Spain under project PID2021-126592OB-C22 and by the European Commission under grant CoreSense (N1.101070254).

\bibliography{sn-bibliography}


\end{document}